\def\paperID{10030}
\def\confName{CVPR}
\def\confYear{2026}

\def\paperTitle{From Illusion to Intention: Visual Rationale Learning for Vision-Language Reasoning}

\def\authorBlock{
$^{1}$Changpeng Wang$^{*}$,\;
$^{2}$Haozhe Wang$^{*}$,\;
$^{3}$Xi Chen,\;
$^{1}$Junhan Liu,\;
$^{4}$Taofeng Xue,\;
$^{4}$Chong Peng,\\
$^{1}$Donglian Qi,\;
$^{2}$Fangzhen Lin,\;
$^{1}$Yunfeng Yan\textsuperscript{\ddag} \\
$^{1}$Zhejiang University \quad $^{2}$The Hong Kong Univerisity of Science and Technology \\
$^{3}$The University of Hong Kong \quad $^{4}$Meituan \\
\textsuperscript{*}Equal contribution\quad
\textsuperscript{\ddag} Corresponding author
}

\newif\ifreview 
\newif\ifarxiv \newcommand{\arxiv}{\arxivtrue}
\newif\ifcamera 
\newif\ifrebuttal 

\arxiv

\pdfoutput=1     
\documentclass[10pt,twocolumn,letterpaper]{article}   
\ifreview \usepackage[review]{cvpr} \fi
\ifarxiv \usepackage[pagenumbers]{cvpr} \fi
\ifrebuttal \usepackage[rebuttal]{cvpr} \fi
\ifcamera \usepackage{cvpr} \fi

\usepackage{graphicx}	
\usepackage{amsmath}	
\usepackage{amssymb}	
\usepackage{booktabs}
\usepackage{times}
\usepackage{microtype}
\usepackage{epsfig}
\usepackage{caption}
\usepackage{float}
\usepackage{placeins}
\usepackage{color, colortbl}
\usepackage{stfloats}
\usepackage{enumitem}
\usepackage{tabularx}
\usepackage{xstring}
\usepackage{multirow}
\usepackage{xspace}
\usepackage{url}
\usepackage{subcaption}
\usepackage{xcolor}
\usepackage[hang,flushmargin]{footmisc}

\ifcamera \usepackage[accsupp]{axessibility} \fi

% ...

\ifarxiv  \fi

\newcommand{\R}[1]{{%
    \textbf{%
        \ifstrequal{#1}{1}{\textcolor{red}{R#1}}{%
        \ifstrequal{#1}{2}{\textcolor{blue}{R#1}}{%
        \ifstrequal{#1}{3}{\textcolor{magenta}{R#1}}{%
        \ifstrequal{#1}{4}{\textcolor{teal}{R#1}}{%
                           \textcolor{cyan}{R#1}%
        }}}}%
    }%
}}

\usepackage{xr-hyper}

\makeatletter
\newcommand*{\addFileDependency}[1]{
  \typeout{(#1)}
  \@addtofilelist{#1}
  \IfFileExists{#1}{}{\typeout{No file #1.}}
}

\makeatother
\newcommand*{\myexternaldocument}[1]{
    \externaldocument{#1}
    \addFileDependency{#1.tex}
    \addFileDependency{#1.aux}
}

\usepackage[most]{tcolorbox}
\tcbset{
  colback=gray!5,  
  colframe=gray!80,  
  boxrule=0.6pt,
  arc=2pt,
  left=6pt,
  right=6pt,
  top=4pt,
  bottom=4pt,
}

\definecolor{cvprblue}{rgb}{0.21,0.49,0.74}
\usepackage[pagebackref,breaklinks,colorlinks,allcolors=cvprblue]{hyperref}
\usepackage[capitalize]{cleveref}
\crefname{section}{Sec.}{Secs.}
\crefname{table}{Table}{Tables}
\crefname{figure}{Fig.}{Figs.}

\ifarxiv \crefname{appendix}{App.}{Apps.}
\else \crefname{appendix}{Suppl.}{Suppls.} \fi

\unless\ifarxiv \myexternaldocument{_supplementary} \fi

\begin{document}
\title{\paperTitle}
\author{\authorBlock}
\maketitle

\begin{abstract}
Recent advances in vision–language reasoning underscore the importance of thinking with images, 
where models actively ground their reasoning in visual evidence. 
Yet, prevailing frameworks treat visual actions as optional tools, boosting metrics but leaving reasoning ungrounded and crops ineffective.
This gap gives rise to the \textbf{illusion of thinking with images}: models seem visually grounded but rely on context-agnostic actions that neither refine perception nor guide reasoning toward correct answers.
We address this problem by reframing visual actions as core reasoning primitives rather than optional tools, which we term visual rationalization, the visual analogue of textual Chain-of-Thought. 
Building on this insight, we propose \textbf{Visual Rationale Learning (ViRL)}, an end-to-end paradigm that grounds training in the visual rationale itself. ViRL integrates (1) Process Supervision with ground-truth rationales, (2) Objective Alignment via step-level reward shaping, and (3) Fine-Grained Credit Assignment to distinguish correct, redundant, and erroneous actions. 
By ensuring each action contributes meaningfully to the reasoning chain, ViRL enables models to ``get the right answer for the right visual reason''.
Trained purely with end-to-end RL, ViRL achieves state-of-the-art results across benchmarks spanning perception, hallucination, and reasoning.
This work establishes visual rationalization as a task-agnostic, process-grounded paradigm for building transparent, verifiable, and trustworthy vision-language models.

\end{abstract}

\section{Introduction}
\label{sec:intro}

\begin{figure}[htbp]
  \centering
  \includegraphics[width=1.0\linewidth]{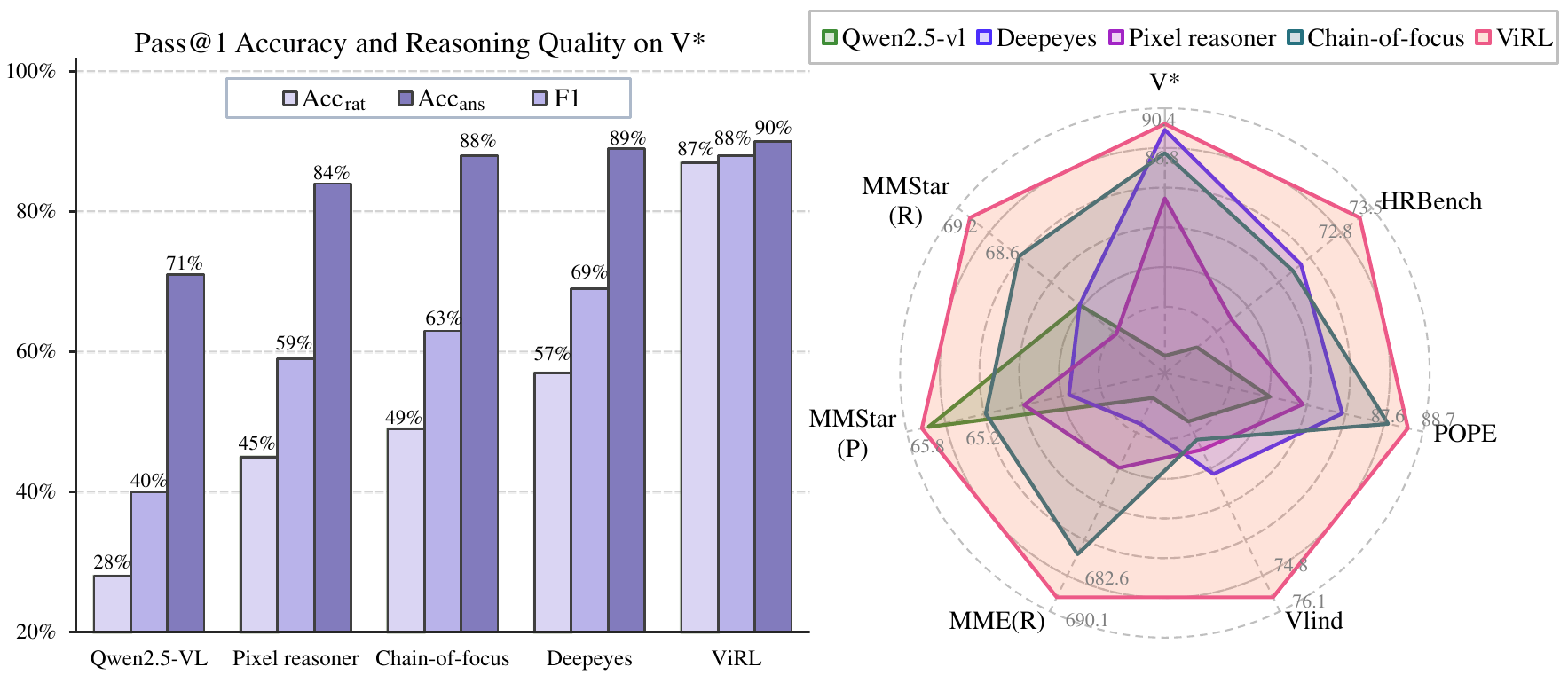}
  \caption{
  Overall comparison of existing visual-reasoning methods and ViRL. \textbf{Left}: answer accuracy($\text{Acc}_{\text{ans}}$), rationale accuracy ($\text{Acc}_{\text{rat}}$), and their joint score ($\mathcal{F}_{1}$) on V*. 
  \textbf{Right}: Radar-chart across multiple reasoning and reliability benchmarks. ViRL improves both rationale fidelity and multi-benchmark robustness, closing the gap between actions and outcomes.
  }
  \label{fig:compare}
\end{figure}

\begin{figure*}[htbp]
  \centering
  \includegraphics[width=1.0\linewidth]{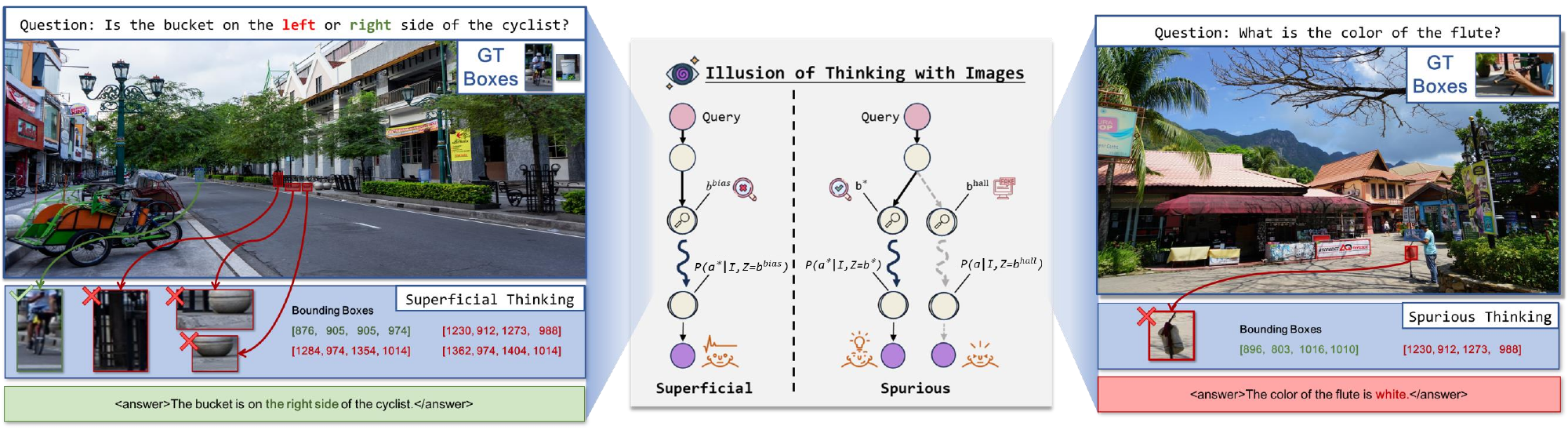}
  \caption{
  Examples illustrating the illusion of ``thinking with images''.
  Even when models execute visual actions, they often zoom into irrelevant regions (superficial) or select misleading cues (spurious), creating the false impression of grounded reasoning while contributing little to answering the question.
  }
  \label{fig:illusion_cases}
\end{figure*}

The advent of Chain-of-Thought (CoT)~\citep{wei2023chainofthoughtpromptingelicitsreasoning} has marked a pivotal moment for large language models, enabling them to solve complex problems by externalizing their reasoning into verifiable, step-by-step textual traces~\cite{verify, wang2025emergent,wang2025reverse}. However, in vision-language settings, reasoning is not confined to language~\citep{LLaVA-OneVision-1.5, , wang2025vl, unimev2,xie2024croc, an2024mc, an2025unictokens, li2025chemvlm}. To solve real-world multimodal tasks~\citep{unicom,lin2025perceive, musellm, Ma_Chen_Zhang_Wu_Ding_2025}, models must not only ``think in text'' but also ``think with images''~\citep{openai2025thinkingwithimages, wang2025pixel}. This paradigm shift moves beyond purely textual reasoning toward reasoning in the pixel space, where models actively interact with the visual world. In particular, zoom-in actions have emerged as a straightforward yet powerful mechanism~\citep{wang2025pixel,wu2024dettoolchainnewpromptingparadigm, hu2024visualsketchpadsketchingvisual, huang2025visualtoolagentvistareinforcementlearning}, allowing models to selectively inspect informative image regions and tightly couple perception with reasoning.

The true potential of ``thinking with images'' lies in its ability to transform a model from a passive observer into an active participant~\citep{wang2025pixel, wang2025code}. By performing zoom-in operations, a model can decompose a complex visual scene into a series of focused inquiries, mimicking the human ability to scrutinize details and gather evidence. This active perception promises to yield models that are: (a) more robust, grounding each step of their reasoning in specific visual evidence rather than relying on spurious correlations; (b) more efficient, by concentrating computational resources on relevant details; and (c) more trustworthy, by producing a transparent and verifiable visual audit trail for their decisions. In this ideal vision, a simple zoom-in is not merely an action but a deliberate, structured component of a causal reasoning chain.

However, a significant gap exists between this ideal and the current reality. As illustrated in Fig.~\ref{fig:illusion_cases}, we identify a fundamental, underexplored problem we term the illusion of ``thinking with images''. This illusion arises when models appear to engage in image-grounded reasoning, yet their visual actions are, in practice, superficial or even spurious, contributing little to the final decision. This issue stems not merely from a choice of learning algorithm, but from a deeper conceptual misunderstanding. The prevailing approach treats visual actions as a form of ``tool use'', where they are considered as an auxiliary mechanism to inspect visual details. This perspective naturally leads to outcome-based supervision, as the zoom-in ``tool'' is only optional for maximizing task success.

The consequence of this view is a stark disconnect between actions and outcomes. As shown in Figure~\ref{fig:compare}, some models achieve high final-answer accuracy (e.g., Deepeyes~\citep{zheng2025deepeyesincentivizingthinkingimages} at 89.1\%) while exhibiting low accuracy in their visual actions (57\%) and performing an excessive number of operations. Such high performance on a benchmark belies a fragile and inefficient process. As shown in our analyses, this illusion results in models that are: (1) \textit{brittle}, as their shortcut behaviors collapse under distribution shifts; (2) \textit{inefficient}, inflating inference costs with redundant actions; and (3) \textit{untrustworthy}, as their spurious reasoning traces undermine interpretability in high-stakes applications.

To break this illusion, we must move beyond the ``tool-use'' metaphor. The sequence of zoom-in operations is not an auxiliary procedure but the reasoning process itself. We clarify and formalize this concept as Visual Rationalization—the visual analogue of textual CoT. In this framework, the reasoning unfolds through an explicit sequence of zoom-in operations that progressively highlight the evidence underlying each conclusion. This reframing elevates zoom-in operations from simple tools to the core components of transparent and verifiable rationales, shifting the learning paradigm from outcome-centric to process-grounded learning and ensuring that models ``get the right answer for the right reason''.

To realize this, we introduce \textbf{Visual Rationale Learning (ViRL)}, an end-to-end learning paradigm that explicitly optimizes the fidelity and utility of the visual reasoning process. A primary obstacle to learning such rationales has been the absence of process-level supervision. Our first key contribution is a process-grounded dataset, which provides ground-truth visual rationales that serve as the evidence foundation for each task. Beyond supervision, this dataset also employs a reasoning-centric filtering strategy to retain tasks that genuinely require image-grounded reasoning. 
With this data, ViRL implements two novel learning mechanisms:
\begin{itemize}[leftmargin=1.5em]
    \item Rationale Fidelity Reward, which measures the fidelity of the visual rationale by directly evaluating how closely the model's sequence of visual foci (i.e., zoom-in coordinates) matches the ground-truth evidence;
    \item Rationale Utility Shaping, a mechanism that shapes the model's policy by defining the utility of each zoom-in action (e.g., correct, redundant, or erroneous), teaching the model to make more deliberate and efficient choices.
\end{itemize}

By directly optimizing the reasoning process, ViRL not only sets a new state-of-the-art on key benchmarks but also produces models with demonstrably superior accuracy and efficiency in their visual rationales. This work establishes a new path for building verifiable and trustworthy vision-language models, demonstrating that to truly think with images, an agent must ground its reasoning in coherent and causally meaningful visual evidence.

\section{Related Work}
\label{sec:related}



\textbf{Large Visual Reasoning Models.}
Recently, many reasoning models have been proposed to handle more complex tasks due to their strong ability to decompose these difficult questions into simple, small queries. Early methods rely only on text to decompose these complex tasks and do not analyze the full inputs~\citep{zhang2025critic, li2025camaenhancingmultimodalincontext,li2025taco, li2025miv}. Like Visual cot \cite{shao2024visualcotadvancingmultimodal}, VPD \cite{hu2023visual}, V* \cite{wu2023vguidedvisualsearch}, Inght-V \cite{dong2025insightvexploringlongchainvisual}, and Llava-cot \cite{xu2025llavacotletvisionlanguage}. These methods decompose the visual questions into sample queries in Chain-of-Thoughts(CoTs). However, they tend to ignore the visual clues when thinking in deep chains. In the latter, many methods introduce visual tools to think with images rather than text only. Visual Sketchpad \cite{hu2024visualsketchpadsketchingvisual}, DetToolChain \cite{wu2024dettoolchainnewpromptingparadigm}, Cropper \cite{lee2025croppervisionlanguagemodelimage}, toolformer \cite{schick2023toolformerlanguagemodelsteach}, AutoCode~\citep{wang2025code}, and react \cite{yao2023reactsynergizingreasoningacting} understand complex tasks and call visual tools like zoom-in, drawing line, and depth perception \cite{chen2025sifthinkerspatiallyawareimagefocus} as additional auxiliary tips through in-context learning without training. However, these methods are mainly made up of visual workflows and cannot truly understand the tools. Later, Models like o3 \cite{openai2025thinkingwithimages}, Pixel Reasoner \cite{wang2025pixel}, Deepeyes \cite{zheng2025deepeyesincentivizingthinkingimages}, Chain-of-focus \cite{zhang2025chainoffocusadaptivevisualsearch}, and OpenThinkIMG \cite{su2025openthinkimglearningthinkimages} demonstrate the ability by dynamically applying visual tools to reasoning in multimodal chains of thoughts~\citep{xia2024mmed,tongyidr,geng2025webwatcher}. 
While these methods improve visual reasoning, their outcome-supervised rewards make the visual thinking process error-prone and susceptible to hallucination. 

\begin{figure*}[htbp]
  \centering
  \includegraphics[width=1.0\linewidth]{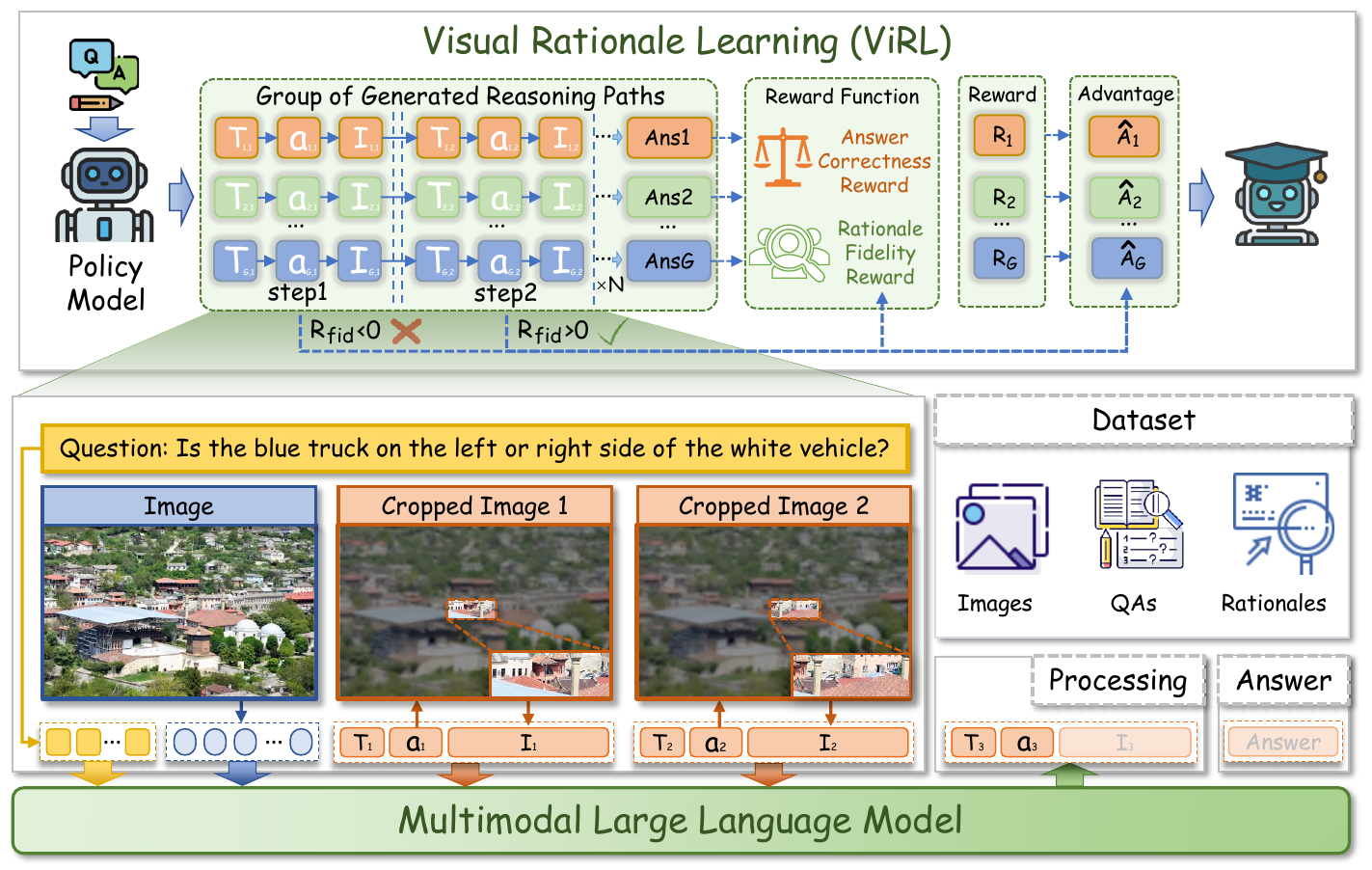}
  \vspace{-5pt}
  \caption{
  The end-to-end visual rationale learning framework. The framework treats zoom-in actions as core reasoning primitives, combining multi-turn learning with process supervision to produce interpretable and verifiable visual reasoning chains.
  }
  \label{fig:framework}
\end{figure*}

\vspace{4pt}
\noindent
\textbf{Reinforcement Learning for MLLM Reasoning.}
Reinforcement Learning, as a technique in the post-training of MLLMs~\cite{deepseekai2025deepseekr1incentivizingreasoningcapability, wang2025vl, liu2024improvedbaselinesvisualinstruction, wang2020learning}, has been shown to effectively enhance reasoning ability in visual reasoning tasks. In particular, Proximal Policy Optimization (PPO) \cite{schulman2017proximalpolicyoptimizationalgorithms} has become the most widely adopted algorithm for aligning multimodal reasoning behaviors with reward signals and has been extensively applied in recent vision-language alignment frameworks \cite{openai2024openaio1card}. More recently, GRPO \cite{deepseekai2025deepseekr1incentivizingreasoningcapability} has been introduced as an efficient alternative, offering faster convergence and improved sample efficiency in complex multimodal reasoning scenarios. Later, to alleviate the vanishing advantages problem~\citep{wang2025vl} during RL training, SSR~\cite{wang2025vl} and DAPO \cite{yu2025dapoopensourcellmreinforcement} was proposed. Several works have further refined the optimization process by either redesigning the reward function or adjusting the advantage estimation. VLM-R1  \cite{shen2025vlmr1stablegeneralizabler1style} incorporates multi-component rewards, such as accuracy, format, and REC, to stabilize training. 
HICRA~\cite{wang2025emergent} discovered emergent reasoning hierarchy in reasoning LLMs and propose to modulate the advantages on critical planning tokens, leading to better strategic exploration and training results. 
In this paper, we focus on shaping reward and credit assignment to ensure that models not only reach correct answers but also ground their reasoning in coherent visual rationales.


\section{Method}
\label{sec:method}
\subsection{Problem Statement}
We first formalize the challenge of ``thinking with images'' by identifying the core constraint of standard Vision-Language Models (VLMs): inherent partial observability. When a VLM first processes an image, its visual encoder creates an information bottleneck. Constrained by finite resolution, this encoding obscures or loses the fine-grained details crucial for complex reasoning. The model, therefore, operates on an incomplete belief state, not the full image.

To overcome this, the model must transform from a passive observer into an active reasoner that sequentially gathers new evidence to resolve uncertainty. This transforms the task into a sequential decision-making problem. We define an action space that allows the model to interleave internal logic with active perception. At any step, the policy $\pi_{\theta}$ can choose from two distinct actions:
\begin{itemize}
    \item \textbf{Textual Rationalization.} The model generates text to construct hypotheses, devise next-step strategies, and integrate multimodal evidence into coherent reasoning.
    \item \textbf{Visual Rationalization.} The model executes a visual operation (such as \emph{zoom-in} on region $b_k$) to actively probe the image and acquire new, high-fidelity information. As noted in our introduction, this is the visual analogue of textual Chain-of-Thought.
\end{itemize}
Our goal is to learn a policy $\pi_{\theta}$ that generates an optimal sequence of these textual and visual rationales.
A naive solution would employ reinforcement learning (RL) using an outcome-only reward, 
to activate reasoning ability. 
The core deficiency of this signal is that it is fundamentally indiscriminate. It cannot distinguish between a correct answer achieved via a lucky guess and one derived from a faithful, step-by-step visual grounding process. Worse, it often creates a pathological local optimum: a policy may discover that exploiting spurious correlations or ``hallucinating'' a plausible-sounding answer is a lower-cost, more facile strategy for obtaining a reward than undertaking a complex, multi-step visual rationalization process that carries a higher risk of failure. 
The outcome-only signal, therefore, can perversely discourage the very reasoning we aim to elicit.

\subsection{Visual Rationale Learning}
To break this illusion, we shift the training objective from outcome accuracy to process-level fidelity and utility, ensuring that each zoom-in action contributes meaningfully to the reasoning trajectory. 
We introduce Visual Rationale Learning (ViRL), a principled framework that learns a policy by explicitly rewarding the reasoning process. ViRL is built on two insightful components designed to solve the challenges of hallucinated rationales and coarse credit assignment:
\vspace{2pt}
\begin{itemize}
  \setlength{\itemsep}{3pt}
  \setlength{\parskip}{0pt}
  \setlength{\parsep}{0pt}
  \item \textbf{Rationale Fidelity Reward} that inextricably links the quality of the reasoning process to its ultimate goal.
  \item \textbf{Fine-Grained Credit Assignment} that identifies the precise contribution of each heterogeneous rationale.
\end{itemize}

\vspace{4pt}
\noindent
\textbf{Rationale Fidelity Reward.} 
Our objective is to learn a rational policy that maximizes task utility while maintaining faithful and concise visual reasoning. 
Given a query $\mathcal{Q}$ and trajectory $\tau$, we optimize
\begin{equation} 
\label{eq:objective} 
\max_{\phi}\; \mathbb{E}_{\tau \sim \pi_\phi(\cdot\mid\mathcal{Q})}\!\big[\mathcal{R}_{\mathrm{total}}(\tau)\big]
\end{equation}
where the total reward jointly evaluates answer correctness, format compliance, and process fidelity:
\begin{equation}
\label{eq:total_reward_repeat}
\mathcal{R}_{\mathrm{total}} 
= \mathcal{R}_{\mathrm{acc}} 
+ \mathcal{R}_{\mathrm{fmt}} 
+ \bar{\mathcal{R}}_{\mathrm{fid}}
\end{equation}
This formulation emphasizes process-level supervision, where each visual rationale directly supports the model’s perceptual and inferential steps.
Each rationalization step $a_k$ receives a fidelity reward based on its spatial alignment 
$u_k=\operatorname{IoU}(b_k,b_k^*)$ 
with the reference rationale $b_k^*$. 

Specifically, it comprises a signed correctness term that directly reinforces plausible reasoning while penalizing misaligned ones, and a discrete refinement bonus activated beyond a soft threshold $h_0$, which further encourages precise spatial alignment as IoU increases. Formally,
\begin{equation}
\label{eq:fidelity}
\mathcal{R}_{\mathrm{fid}}(a_k)
= \mathcal{R}_{\mathrm{base}} \cdot \mathrm{sign}(u_k - h_0)
+ \eta\Big\lfloor \frac{\max(0,u_k - h_0)}{\Delta h} \Big\rfloor
\end{equation}
where $\mathrm{sign}(u_k - h_0)$ provides the signed correctness signal, $\eta$ controls the magnitude of the refinement bonus, and $\lfloor \cdot \rfloor$ denotes the discretization operator that introduces stepwise gains. Specifically, $\lfloor (u_k - h_0)/\Delta h \rfloor$ increases by one for every $\Delta h$ improvement in IoU, thereby modeling progressive alignment refinements beyond the soft threshold $h_0$.
If no visual action is taken, $\mathcal{R}_{\mathrm{fid}}(a_k)=0$.

Aggregating the step-level rewards across the reasoning trajectory yields the overall fidelity score:
\begin{equation}
\label{eq:traj_fidelity}
\bar{\mathcal{R}}_{\mathrm{fid}}
= \frac{1}{N}\sum_{k=1}^{N}\big[\mathcal{R}_{\mathrm{fid}}(a_k) - \rho(C_k)\big]
\end{equation}
where averaging ensures fair credit assignment across variable-length trajectories. 
A smooth redundancy penalty $\rho(C_k)$ grows beyond a soft budget, discouraging repetitive or overlapping rationales and promoting concise reasoning.

\vspace{4pt}
\noindent
\textbf{Fine-Grained Credit Assignment.}
With the trajectory reward $R(\tau)$, we confront the second challenge—the credit assignment problem, which is further exacerbated by the \textit{heterogeneous} nature of our action space. 
A single trajectory-level advantage is too coarse, indiscriminately crediting or blaming all steps and encouraging erroneous visual rationales within successful trajectories.
We solve this with a bi-level credit assignment strategy that modulates the advantage signal based on rationale-specific quality.

\begin{itemize}
\item \textbf{Trajectory-Level Advantage:} First, we compute a coarse advantage signal for the entire trajectory $\tau_i$ by comparing its reward to the average of a group of $G$ trajectories \cite{shao2024deepseekmathpushinglimitsmathematical}:
\begin{equation}
A_i = R(\tau_i) - \frac{1}{G} \sum_{j=1}^{G} R(\tau_j)
\label{eq:traj_advantage}
\end{equation}
This indicates whether the trajectory is globally advantageous ($A_i > 0$) or disadvantageous ($A_i < 0$).

\item \textbf{Rationale-Level Adjustment:} To obtain fine-grained credit assignment, we differentiate the coarse advantage signal. Textual rationales preserve the original advantage $A_i$, whereas \textit{visual rationales} are adaptively modulated by their fidelity (Eq.~\ref{eq:fidelity}). The resulting adjusted advantage for each action $a_t$ is denoted as $\hat{A}_{i,t}$:
\begin{equation}
\hat{A}_{i,t} = A_i \cdot h(a_t)
\label{eq:adj_advantage}
\end{equation}
\end{itemize}
where $h(a_t)$ is a simple modulator based on the rationale:
\begin{equation}
h(a_t) = 
\begin{cases}
    h_{\text{good}}, & \text{if } a_t \text{ is Good Visual Rationale } (\mathcal{R}_{\text{fid}} > 0) \\
    h_{\text{bad}},  & \text{if } a_t \text{ is Bad Visual Rationale } (\mathcal{R}_{\text{fid}} \le 0) \\
    1,          & \text{if } a_t \text{ is Textual Rationale}
\end{cases}
\label{eq:modulator}
\end{equation}
This logic provides a highly principled teaching signal:
\begin{itemize}
    \item \textbf{In an advantageous Trajectory ($A_i > 0$):} the model amplifies credit for high-fidelity visual rationales ($h_{\text{good}} > 1$) while attenuating it for low-fidelity ones ($h_{\text{bad}} < 1$). This guides the policy to attribute success \textit{to} its precise and faithful visual reasoning.
    
    \item \textbf{In a disadvantageous Trajectory ($A_i < 0$):} the model amplifies blame for bad visual rationales ($h_{\text{bad}} > 1$) and mitigate blame for any good ones ($h_{\text{good}} < 1$). This guides the policy to recognize that failure stems from inaccurate or misleading visual reasoning.
\end{itemize}

\vspace{4pt}
\noindent
\textbf{Policy Optimization.}
This fine-grained advantage $\hat{A}_{i,t}$ is then used to update the policy $\pi_{\theta}$ via a standard PPO-style objective. This bi-level ViRL framework compels the policy to learn the true contribution of each visual rationale, moving the model \textit{from illusion to intention} by ensuring it generates reasoning that is not just faithful, but also utility-enhancing.

\begin{figure}[t!]
  \centering
  \includegraphics[width=\linewidth]{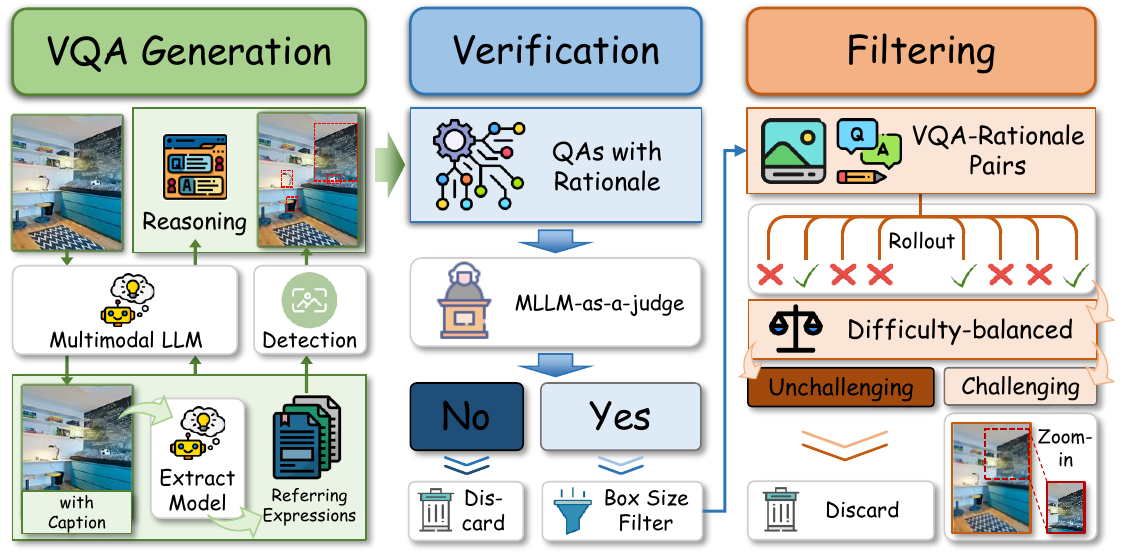}
  \caption{A three-stage pipeline for process-grounded data curation, involving generation, verification, and reasoning-centric filtering. 
  }
  \label{fig:sample}
\end{figure}

\subsection{Process-Grounded Dataset Generation Pipeline} 
As illustrated in \autoref{fig:sample}, the dataset is constructed through a three-stage pipeline comprising generation, verification, and filtering. The pipeline ensures both high-quality reasoning data and explicit process-level supervision, forming the foundation of the visual rationale learning paradigm.

\textbf{(1) Generation.} 
We construct reasoning-oriented VQA supervision by leveraging region captions from GRIT~\citep{wu2022gritgenerativeregiontotexttransformer}. For each region, the model extracts referring expressions and generates questions that require implicit visual reasoning rather than explicit object naming. The corresponding visual rationale is obtained by localizing the referring expression with a detector and padding the predicted box to capture necessary context. To maintain region diversity, we apply NMS to filter overlapping candidates.
This process yields {question, answer, rationale region} triplets that serve as supervision for training visual reasoning behaviors.

\textbf{(2) Verification.} 
The generated VQA–rationale pairs are subsequently verified via MLLM-based consistency checks, evaluating (i) whether the answer is correct with respect to the visual input, and (ii) whether the annotated rationale aligns with the reasoning target implied by the question. Each sample is re-evaluated using the hinted rationale region and corresponding QA pair, and only those satisfying both semantic correctness and rationale consistency are preserved.

\textbf{(3) Reasoning-Centric Filtering.} 
Finally, we apply a reasoning-centric filtering stage to ensure that the remaining samples genuinely require image-grounded reasoning. First, samples with overly large rationale regions are removed, as such tasks can typically be solved without localized visual evidence. Second, we conduct an offline difficulty balancing by rolling out multiple reasoning attempts: questions solvable without visual grounding are down-weighted, while those that critically depend on localized cues are retained. This process yields a dataset that encourages models to think with images, providing reliable step-level visual rationales for process-grounded learning.

\section{Experiments}
\label{sec:experiments}

\begin{table*}[t]
\centering
\caption{Our main results on the six evaluated benchmarks. MMStar$^{\dagger}$ evaluates four core categories (coarse perception, fine-grained perception, instance reasoning, and logical reasoning), focusing on general perception-oriented reasoning.}
\label{tab:main_results}
\begin{tabular}{lccccccc}
\toprule
\multirow{2}{*}{Model} & \multirow{2}{*}{Size} &
\multicolumn{2}{c}{\textbf{Perception-Oriented}} &
\multicolumn{2}{c}{\textbf{Reliability-Oriented}} &
\multicolumn{2}{c}{\textbf{Reasoning-Oriented}} \\
\cmidrule(lr){3-4} \cmidrule(lr){5-6} \cmidrule(lr){7-8}
 & & V* & HRBench-4K & POPE & VLind & MME(R) & MMStar$^{\dagger}$ \\
\midrule
\multicolumn{8}{l}{\textit{Models w/o ``Thinking with images''}} \\
\midrule
GPT-4o               & -   & 45.0 & 46.8 & 84.6 & \textbf{89.8} & 674.6 & \textbf{73.0} \\
GPT4o-mini           & -   & 50.8 & 48.0 & 83.3 & -- & 564.3 & -- \\
LLaVA-OV             & 7B  & 72.8 & 64.7 & 88.3 & 54.8 & 415.4 & 62.1 \\
Qwen2.5-VL           & 7B  & 71.4 & 69.2 & 86.6 & 67.4 & 610.1 & 66.6 \\
Qwen2.5-VL           & 32B & 81.2 & 73.4 & 85.7 & 81.2 & 645.6 & 68.8 \\
\midrule     
\multicolumn{8}{l}{\textit{Models with ``Thinking with images''}} \\
\midrule
Visual Sketchpad (GPT-4o) & -   & 80.4 & -- & -- & -- & -- & -- \\
SEAL                      & 7B  & 75.4 & -- & -- & -- & -- & -- \\
DyFo                      & 7B  & 81.2 & -- & -- & -- & -- & -- \\
Deepeyes*                  & 7B  & 88.9 & 73.1 & 87.7 & 70.0 & 620.7 & 65.4 \\
Chain-of-focus            & 7B  & 88.0 & 70.5 & 88.4 & 68.3 & 673.5 & 66.6 \\
Pixel reasoner*            & 7B  & 84.3 & 72.8 & 87.1 & 68.8 & 638.5 & 64.8 \\
\midrule
\multicolumn{8}{l}{\textit{Ours (Initialized from Qwen2.5-VL-7B)}} \\
\midrule
\textbf{ViRL}             & 7B  & \textbf{90.1} & \textbf{75.3} & \textbf{88.7} & \textbf{76.1} & \textbf{691.0} & \textbf{67.5} \\
\bottomrule
\end{tabular}

\end{table*}

\begin{figure*}[htbp]
  \centering
  \includegraphics[width=1.0\linewidth]{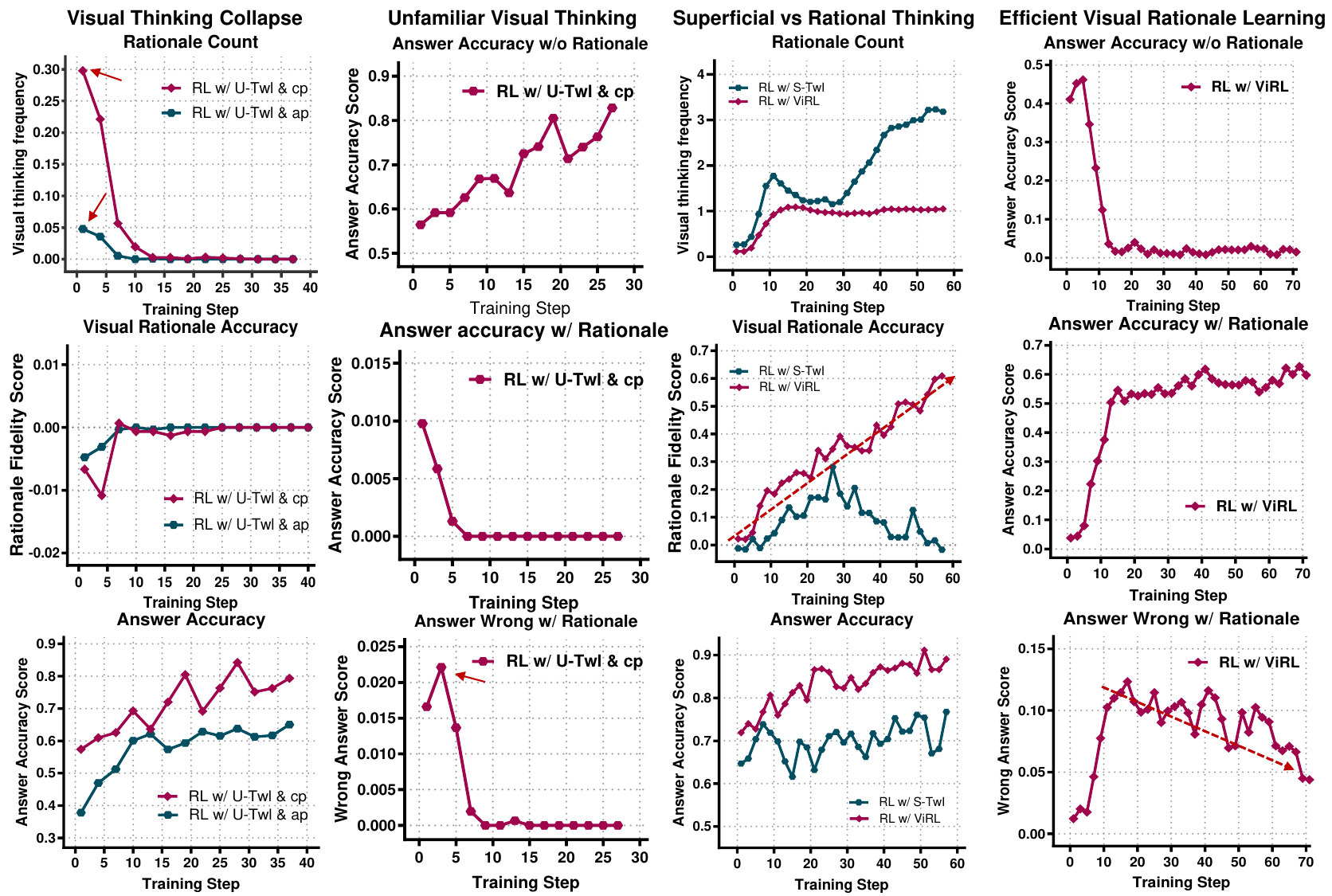}
  \caption{Training dynamics of visual rationale learning.
From left to right: (a) collapse of visual thinking, (b) the underlying cause: unfamiliarity with visual thinking, (c) 
superficial vs. ViRL-grounded reasoning,
and (d) the emergence of faithful reasoning under our ViRL framework.
U-TwI and S-TwI denote outcome-only and naive step-wise reward baselines.
cp / ap represent clear / ambiguous prompts.}
  \label{fig:inght1}
\end{figure*}

\subsection{Experimental Setup}
We construct a 200k-scale dataset by collecting samples from Visual-Cot \cite{shao2024visualcotadvancingmultimodal}, GQA \cite{hudson2019gqanewdatasetrealworld}, TextVQA \cite{singh2019vqamodelsread}, Visual7W \cite{zhu2016visual7wgroundedquestionanswering}, InfographicsVQA \cite{mathew2021infographicvqa}, and GRIT \cite{wu2022gritgenerativeregiontotexttransformer}, covering general understanding, visual grounding, reasoning, and OCR tasks. 
After processing within the data generation pipeline (Sec .~3.2), 8k high-quality samples are retained. Since all data have open-vocabulary answers that are difficult to evaluate, we repackage them as multiple-choice questions by synthesizing 3–7 plausible distractors from the VQA and global descriptions to preserve task difficulty. 
We fine-tune Qwen2.5-VL-7B \cite{bai2025qwen25vltechnicalreport} on 8 A100 GPUs for 80 iterations. Following DAPO \cite{yu2025dapoopensourcellmreinforcement}, we adopt clip-higher, dynamic sampling, and token-level policy loss to enhance data utilization. Training uses a batch size of 64 with 16 rollouts per prompt, allowing up to 6 rounds of cropping. The learning rate is $1\times 10^{-6}$, the maximum response length is 20,480 tokens, and neither KL nor entropy regularization is applied.

\subsection{Baselines and Benchmarks.}
We conduct comprehensive evaluations against a wide range of competitive baselines. For closed-source models, including GPT-4o \cite{openai2024gpt4technicalreport} and GPT4o-mini \cite{openai2024openaio1card}. For open-source models, we benchmark against state-of-the-art visual language models such as Qwen2.5-VL \cite{bai2025qwen25vltechnicalreport} and LLaVA-OneVision \cite{li2024llavaonevisioneasyvisualtask}. We also compare with workflow-based methods like Visual Sketchpad~\cite{hu2024visualsketchpadsketchingvisual}, SEAL~\cite{wu2023vguidedvisualsearch}, and DyFo~\cite{li2025dyfotrainingfreedynamicfocus}, which design explicit multi-step reasoning pipelines. In addition, we evaluate against recent ``thinking with images'' approaches, including Deepeyes \cite{zheng2025deepeyesincentivizingthinkingimages}, Chain-of-focus \cite{zhang2025chainoffocusadaptivevisualsearch}, and Pixel Reasoner \cite{wang2025pixel}. 
For evaluation, we adopt a comprehensive benchmark suite spanning three complementary dimensions of vision–language reasoning: (i) Perception-Oriented benchmarks, including fine-grained perception (V*) \cite{wu2023vguidedvisualsearch} and high-resolution perception (HRBench) \cite{wang2024divideconquercombinetrainingfree}; (ii) Reliability-Oriented benchmarks, covering hallucination tendency (POPE) \cite{li2023evaluatingobjecthallucinationlarge} and language-prior reliance (VLind) \cite{lee2024vlind}; 
and (iii) Reasoning-Oriented benchmarks, assessing general multimodal reasoning capabilities (MME(R) \cite{fu2024mmecomprehensiveevaluationbenchmark} and MMStar \cite{chen2024we}). 
This comprehensive design ensures a fair and multi-dimensional comparison.
In addition to external benchmarks, we further report two internal diagnostics that quantify how models engage in image-grounded reasoning: Rationale Accuracy, which measures whether visual zoom-ins successfully hit the ground-truth evidence, and Rationale Count, which captures the frequency of visual-thinking steps. We also provide a joint $\mathcal{F}_{1}$ score that summarizes answer correctness and rationale fidelity. Complete definitions of these metrics are included in Appendix~A.

\subsection{Main Results}
\textbf{Perception-Oriented Results.}
As shown in the Perception-Oriented results of Table~\ref{tab:main_results}, fine-grained and high-resolution perception remain core challenges for vision-language models. Model scaling offers limited benefits (Qwen2.5-VL-7B vs. 32B), while introducing explicit visual reasoning (e.g., Visual Sketchpad) yields notable gains on V* and HRBench.
\textbf{ViRL} achieves state-of-the-art performance (90.1 on V*, 75.3 on HRBench) by decomposing complex scenes via targeted zoom-in operations, enabling selective and interpretable attention to informative regions.
Compared to answer-grounded approaches(+6.1 over Pixel reasoner on V*), ViRL demonstrates stronger fine-grained perception,
attributed to its supervision on intermediate visual rationales.

\noindent \textbf{Reliability-Oriented Results.}
As shown in the Reliability-Oriented results of Table~\ref{tab:main_results},
ViRL achieves the highest POPE score, indicating strong resistance to hallucination. Notably, ViRL also attains the best performance on VLind, reflecting its reduced reliance on language priors. This is not merely a byproduct of lower hallucination rates: VLind explicitly probes whether models actively ground their predictions in visual evidence. 
Higher VLind scores reflect that the model actively relies on image-grounded cues to override textual regularities based on language priors. 
ViRL’s superior performance (+8.7 compared to baseline) demonstrates that its active visual reasoning effectively mitigates language-prior–induced hallucinations, confirming that the model genuinely leverages visual evidence to guide its predictions.


\noindent \textbf{Reasoning-Oriented Results.}
As shown in the Reasoning-Oriented results of Table~\ref{tab:main_results}, \textbf{ViRL} achieves the highest performance (691.0 on MME(R) and 67.5 on MMStar$^{\dagger}$), outperforming both proprietary and open-source models.
Unlike prior methods that employ tools as auxiliary aids with limited gains (e.g., Deepeyes reports reductions of 10 on MME(R) and 1.2 on MMStar$^{\dagger}$), ViRL instead elevates the zoom-in operation into a core reasoning primitive, ensuring that each visual action contributes meaningfully to the inference trajectory and remains both informative and image-grounded.
This process-level supervision yields more stable perception, coherent multi-step inference, and superior performance on complex reasoning tasks, demonstrating that faithful and interpretable reasoning naturally emerges when perception and reasoning are trained as a unified, evidence-driven system.

\begin{table}[t]
\centering
\small
\setlength{\tabcolsep}{6pt}
\renewcommand{\arraystretch}{1.1}
\caption{
Ablation results on \textbf{V*}, reporting answer accuracy $\text{Acc}_{\text{ans}}$, rationale accuracy $\text{Acc}_{\text{rat}}$, rationale count $\mathcal{C}_{\text{rat}}$, and a joint $\mathcal{F}_{1}$ score.
}
\label{tab:ablation_combined}
\begin{tabular}{lcccc}
\toprule
Method & $\text{Acc}_{\text{ans}}$ & $\text{Acc}_{\text{rat}}$ & $\mathcal{C}_{\text{rat}}$ & $\mathcal{F}_{1}$ \\
\midrule
\multicolumn{5}{l}{\textit{Component Ablations}} \\
\midrule
Base model                        & 71.4 & 28.2 & 2.65 & 0.40 \\
RL w/o Rationale Fidelity         & 87.6 & --   & --   & --   \\
RL w/o VTC Data                   & 79.9 & 47.3 & 1.39 & 0.59 \\
RL w/o FGCA                      & 88.9 & 78.2 & 1.17 & 0.83 \\
\midrule
\multicolumn{5}{l}{\textit{Rationale Fidelity Reward Variants}} \\
\midrule
+ Rationale (low) & 87.2 & 75.5 & 1.12 & 0.75 \\
+ Rationale (high) & 87.9 & - & - & - \\
\midrule
\textbf{ViRL (Ours)}              & \textbf{90.4} & \textbf{87.3} & \textbf{1.04} & \textbf{0.88} \\
\bottomrule
\end{tabular}
\end{table}

\subsection{Ablation Study}

\vspace{4pt}
\noindent
\textbf{Does process supervision matter?}
As shown in Table~\ref{tab:ablation_combined},
removing the rationale fidelity reward (answer-only supervision) collapses the model’s ability to think with images.
Although answer accuracy remains superficially strong (87.6\%), the absence of explicit visual rationale signals leads to the disappearance of intermediate visual reasoning traces.
This reveals that under pure outcome supervision, the model simply learns to ``answer in context'' by exploiting latent correlations in the prompt and visual content, bypassing the need for actual visual rationale invocation.
To probe this effect, we analyze how different reward designs shape behavior.
A high success threshold ($h_0=0.5$) over-penalizes early failures, saturating feedback with negatives and reverting the model to answer-only behavior.
A low threshold ($h_0=0.2$) over-rewards weak alignments, blurring quality distinctions and inducing redundant, noisy rationales, reducing success to 75.5\%.
Our proposed method balances this trade-off by adapting reward granularity. 
Lenient early to foster exploration, then progressively stricter as competence grows, which yields stable rationale learning trajectories.

\vspace{4pt}
\noindent
\textbf{Is visual thinking-centric data necessary?}
As shown in Table~\ref{tab:ablation_combined}, removing visual thinking–centric data (RL w/o VTC Data) and substituting unfiltered samples markedly undermines supervision fidelity. Two characteristic failure modes emerge. First, Visual rationale activations become frequently misaligned with the image, injecting noisy cognitive signals that destabilize perceptual grounding and collapse rationale accuracy to 47.3\%. 
Second, the absence of reasoning-centric verification admits trivial or ambiguous instances that require no localized evidence.
Consequently, answer accuracy declines to 79.9\%, while visual rationales become increasingly frequent yet poorly grounded, revealing blurred perceptual principles and unstable exploratory behavior.

\vspace{4pt}
\noindent
\textbf{Can the model differentiate superficial rationales?}
As shown in table~\ref{tab:ablation_combined}, removing fine-grained credit assignment (RL w/o FGCA) exposes the limitation of uniform crediting.
This indiscriminate assignment undermines the model’s ability to prioritize meaningful evidence, yielding only moderate rationale accuracy.
In contrast, the full ViRL framework restores selective crediting, achieving the best balance ($\mathcal{F}_{1}= 0.88$) and enabling robust, scalable visual reasoning.

\subsection{Further analysis}

\paragraph{From Latent Ability to Structured Reasoning.}
As shown in Fig.~\ref{fig:inght1}a, we observed a characteristic: an initial spike in zoom-in invocations followed by a steep decay to near-zero invocation frequency, even as answer accuracy remains non-trivial. 
We term this behavior \textbf{visual thinking collapse}. 
Our empirical study points to three mechanistic contributors:
\begin{itemize}[leftmargin=1.5em]
\item \textbf{Prompt Effects.} 
As shown by the ``Rationale Count'' in Fig.~\ref{fig:inght1}a, 
Clear prompts more strongly tie zoom-ins to reasoning, fostering stronger early exploration, 
whereas prior designs treat visual thinking as a vague action and yield weak early engagement. (Appendix~C)

\item \textbf{Unfamiliarity with Visual Thinking.} 
Early zoom-ins often inject misleading regional evidence, yielding low-utility explorations and increasing ``Answer Wrong w/ Rationale'' errors (Fig.~\ref{fig:inght1}b).

\item \textbf{Outcome-only Supervision.} 
Outcome-only rewards encourage avoiding rationales: accuracy improves without zoom-ins but degrades with them, reinforcing risk-averse, answer-first behavior (Fig.~\ref{fig:inght1}b).

\end{itemize}

\vspace{4pt}
\noindent
\textbf{Invocation Frequency \(\neq\) Invocation Quality.}
A naive fix is to reward zoom-ins, but Fig.~\ref{fig:inght1}c shows that this boosts activity without improving, and often harming task accuracy.
The model learns to invoke for the sake of invoking: rationale counts rise while per-invocation fidelity stagnates. This reveals the illusion of ``thinking with images'': the model appears to be actively engaging tools, yet the underlying reasoning quality stagnates or deteriorates. 
Thus, genuine ``thinking with images'' emerges not from more invocations, but from learning to strategically ground reasoning in the right visual evidence.
Efficient visual reasoning requires \textbf{sparse yet precise} grounding.
As shown in ``Visual Rationale Accuracy'' and ``Answer Wrong w/ Rationale'' (Fig.~\ref{fig:inght1}c,d), high-performing agents exhibit an early increase in invocation frequency during exploration, followed by convergence to a stable regime with markedly higher per-invocation accuracy and fewer illusion-induced errors. This transition reflects a shift from noisy trial-and-error to targeted, high-value visual reasoning. 
In contrast, naive invocation-level incentives lead to persistent over-activation, inflating tool use without improving reasoning quality.

\section{Conclusion}
\label{sec:conclusion}


In this work, we addressed the \textbf{illusion of ``thinking with images''} by reframing visual actions from auxiliary tools into core reasoning primitives. We introduced \textbf{ViRL}, a process-supervised reinforcement learning framework that explicitly optimizes the fidelity and utility of visual reasoning chains. By grounding supervision at the process level rather than solely on final outcomes, ViRL enables models to strategically attend to the right visual evidence at the right time. This shift not only yields state-of-the-art performance across perception, hallucination resistance, and visual reasoning benchmarks, but also establishes a verifiable and interpretable foundation for building trustworthy vision–language agents. Future work will further generalize this paradigm to more complex reasoning modalities and interactive settings.

{\small 
\bibliographystyle{ieeenat_fullname} 
\bibliography{11_references} 
}

\ifarxiv \clearpage \appendix 

\section{Metrics}
To comprehensively evaluate both the accuracy of answers and the quality of visual reasoning behaviors, we introduce three metrics: \textbf{Rationale Accuracy}, \textbf{Rationale Count}, and \textbf{F1 Score}. These metrics are designed to disentangle what the model answers from how it reasons, enabling a more fine-grained understanding of visual thinking performance.

\textbf{Rationale Accuracy.} 
Rationale Acc measures whether the visual rationale invoked during reasoning effectively hits the ground-truth evidence.  
Formally, let $R$ denote the invoked (predicted) rationale region and $G$ the ground-truth bounding box. We define Rationale Accuracy as the fraction of the ground-truth area covered by the invoked region:
\[
\text{Acc}_{\text{rat}}(R,G) \;=\; \frac{\mathrm{Area}(R \cap G)}{\mathrm{Area}(G)}.
\]
We adopt this coverage-based formulation instead of a conventional IoU for two main reasons. 
First, visual thinking (zoom-in) is a reasoning behavior rather than a strict localization task: effective zoom-ins often include contextual regions around the target to capture supporting cues, and such evidence-inclusive regions should not be penalized. 
Second, our objective is to evaluate whether the model accesses the critical evidence needed for inference, not how tightly it localizes the target. 
Measuring coverage relative to the ground-truth region more faithfully reflects this grounding quality.

\textbf{Rationale Count.} 
Rationale Count reflects the average number of visual thinking steps taken per reasoning trajectory. 
It provides insight into how frequently the model engages in image-grounded reasoning when answering questions. 
An excessively low count suggests shortcut behaviors with insufficient visual exploration, while an excessively high count often indicates inefficient or unstable reasoning with redundant visual operations.

\textbf{$\mathcal{F}_{1}$ Score.}
To jointly evaluate perception and reasoning quality, we compute the $\mathcal{F}_{1}$ score as the harmonic mean of Answer Accuracy ($\text{Acc}_{\text{ans}}$) and Rationale Accuracy($\text{Acc}_{\text{rat}}$).  
This formulation emphasizes that strong performance requires both correct answers and correct evidence grounding, rather than excelling in only one aspect:
\[
\mathcal{F}_{1} = 2 \times \frac{\text{Acc}_{\text{ans}} \times \text{Acc}_{\text{rat}}}{\text{Acc}_{\text{ans}} + \text{Acc}_{\text{rat}}}.
\]
A high $\mathcal{F}_{1}$ indicates not only good answer quality but also faithful visual grounding, reflecting more reliable multimodal reasoning.

\begin{tcolorbox}[title={Visual Question Generation Prompt}]
You are a visual question generation expert.\\
Given the detailed visual description of a specific region in an image (below), your task is to create a multiple-choice question based \textbf{only} on that region, but which \textbf{requires implicit reasoning that uses the provided global context}. Do NOT produce questions that simply ask to name or directly identify the object in the region. Instead, design a question that demands reasoning about attributes, affordances, relations, or roles that combine information from the local region and the global scene.

\medskip
Region Description (local evidence):\\
\texttt{""" \{local\_caption\} """}

\medskip
Global Context (global scene information — use this to hint reasoning):\\
\texttt{""" \{global\_caption\} """}

\medskip
Candidate options (only one is correct):\\
\texttt{\{option\_str\}}

\medskip
Guidelines (must follow):\\
1. \textbf{Implicit target}: The question should target an implicit reasoning goal (e.g., function, probable action, relative attribute, relation to other objects, temporal/state inference), not direct object naming.\\
2. \textbf{Use global context}: The correct answer must be inferable only when the local region is interpreted together with the global context. Avoid questions solvable from the local caption alone.\\
3. \textbf{Plausible distractors}: All incorrect options should be plausible given partial evidence, but only one option should be consistent with both local and global cues.\\
4. \textbf{Conciseness \& clarity}: Keep the question concise and specific; do not introduce new visual facts not present in the provided captions.\\
5. \textbf{Format strictly}: Output must follow the exact format below (question text, choices, and final Answer line).

\medskip
\textbf{Expected output format:}\\
\texttt{<question text>?}\\
\texttt{choices:}\\
\texttt{A: ...}\\
\texttt{B: ...}\\
\texttt{C: ...}\\
\texttt{D: ...}\\
\texttt{Answer: \{correct\_answer\}}
\end{tcolorbox}

\begin{tcolorbox}[title={VQA and Grounding Evaluation Prompt}]
You are a visual question answering and grounding evaluation expert.

\medskip
\textbf{Task:}\\
Given a question, its answer, and the corresponding image (with a red bounding box corresponding to the reference area for question–answer pairs), your task is to determine:
\begin{enumerate}
  \item Whether the \textbf{answer is correct} based on the image and question.
  \item Whether the \textbf{red bounding box} matches the question–answer pair, i.e., whether this region contains all the necessary visual evidence to support answering the question correctly.
\end{enumerate}

\medskip
\textbf{Instructions:}\\
- Think step by step, carefully analyzing the question, answer, image, and red bounding box.\\
- \textbf{ONLY} output the final judgment in the exact format: \texttt{\textbackslash boxed\{1\}} if both conditions are true, else \texttt{\textbackslash boxed\{0\}}.

\medskip
\textbf{Input Example:}\\
\texttt{Question: \{question\}}\\
\texttt{Given Answer: \{answer\}}\\
\texttt{Bounding Box Coordinates (padded 10\%): \{padded\_bbox\}}\\
This box has been marked with a red rectangle in the original image.\\
Please pay attention to this area and determine whether it contains the key evidence for answering the question, and whether the given answer is correct.
\end{tcolorbox}

\section{Supplementary Data Explanation}

In this appendix, we provide a brief explanation of the data used for VQA generation and evaluation. Compared with traditional datasets that mainly rely on large bounding boxes and direct object localization, our dataset focuses on \textbf{reasoning-centric} question construction.

Specifically, the designed prompts guide the model to generate questions that require implicit reasoning based on both the local region and the global scene context. This shifts the focus from merely identifying visual objects to understanding their relations, attributes, or roles in the broader scene.

The verification is conducted after generating the VQAs. With the VQA and Grounding Evaluation Prompt, we judge the generated VQA from two aspects:
(1) Answer Correctness: Based on the Question and the Visual input, is the provided answer correct?
(2) Rationale consistency: Consider whether the visual rationale annotations fully cover the relevant object or region required for the reasoning trajectory.

As illustrated in the \autoref{fig:dataset_compare1} and \autoref{fig:dataset_compare2}, traditional data encourages straightforward localization-based answering, while our reasoning-driven data embeds inference cues in the question itself, stimulating deeper visual thinking. This design better aligns with real-world scenarios where answers often depend on implicit reasoning rather than direct perception.

\begin{figure*}[htbp]
  \centering
  \includegraphics[width=1\linewidth]{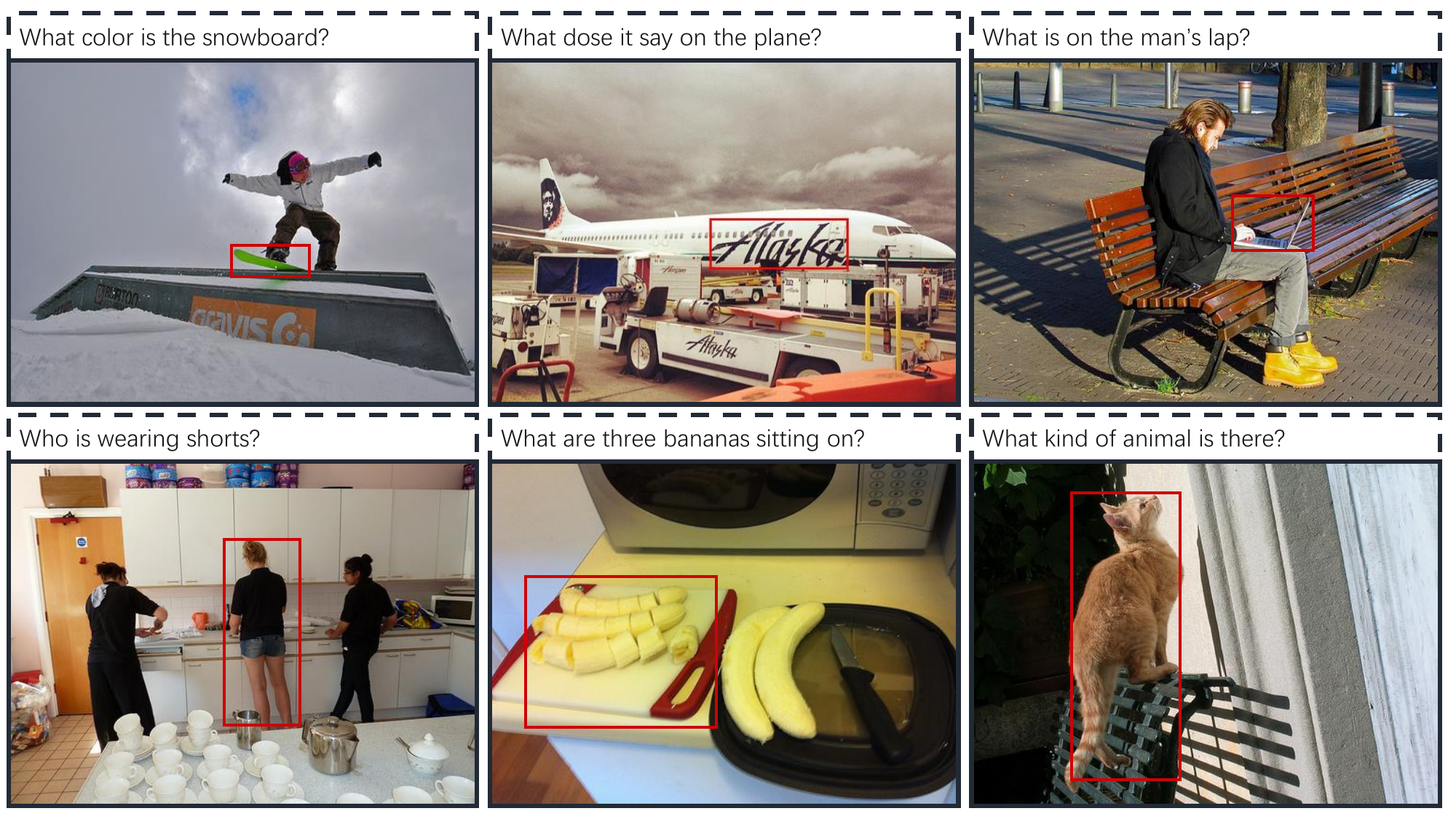}
  \caption{Traditional datasets typically involve large and easily identifiable target regions, which reduces the need for fine-grained visual reasoning and encourages shallow localization behavior rather than active visual thinking.}
  \label{fig:dataset_compare1}
\end{figure*}
\begin{figure*}[htbp]
  \centering
  \includegraphics[width=1\linewidth]{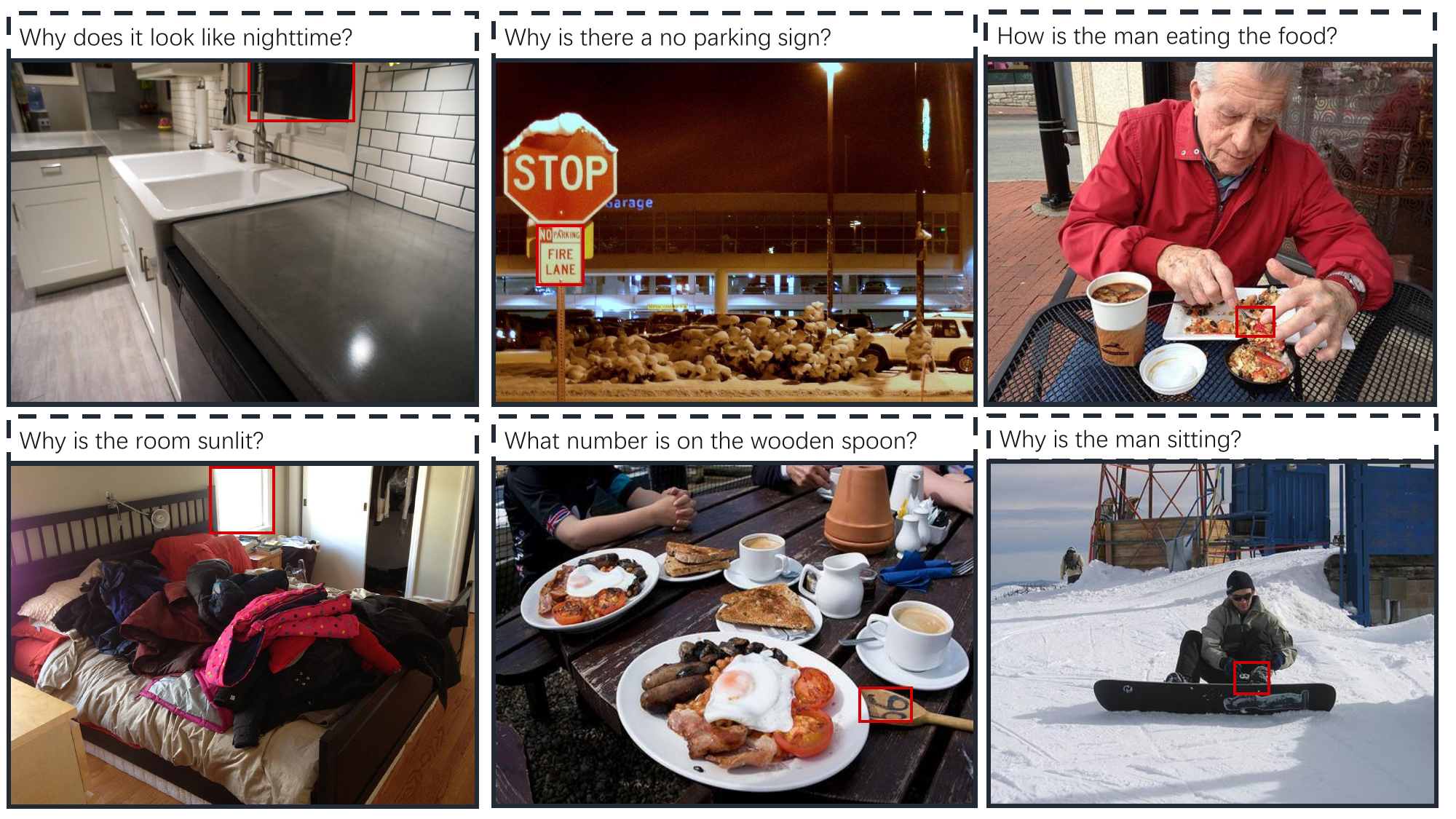}
  \caption{Our dataset emphasizes fine-grained and context-dependent targets, encouraging models to actively engage in visual thinking rather than relying on shallow localization.}
  \label{fig:dataset_compare2}
\end{figure*}

\section{Prompts}

In this section, we provide the prompt designs used for visual reasoning. 
We compare the \textbf{Clear Prompt} with the \textbf{Ambiguous Prompt} commonly adopted in existing works.
Unlike traditional designs that treat \texttt{zoom-in} merely as an optional tool invocation, our prompt explicitly integrates zoom-in as part of the reasoning trajectory. 
This encourages the model to actively gather visual evidence before producing an answer, rather than passively attaching tool calls to the end of reasoning.

\subsection{Ambiguous Prompt}
The ambiguous prompt design gives a brief instruction about zoom-in usage, but does not emphasize its role in reasoning. 
As a result, models may ignore zoom-in actions or invoke them superficially without integrating them into their decision-making process.

\begin{tcolorbox}[colback=gray!5,colframe=gray!50,title=Ambiguous Prompt Words]
\small
Think first, call \textbf{image\_zoom\_in\_tool} if needed, then answer. \\
Format strictly as: \\
\texttt{<think>...</think>} \\
\texttt{<tool\_call>...</tool\_call>} (if tools needed) \\
\texttt{<answer>...</answer>}
\end{tcolorbox}

\begin{tcolorbox}[colback=gray!5,colframe=black,title=Clear Prompt Words (Ours)]
\small
Think carefully before answering. \\
If the current view is not enough (e.g., the image is too large/unclear/complicated) or if a previous zoom-in was incorrect, call \textbf{image\_zoom\_in} to gather better evidence and do not give \texttt{<answer>} in the same step. \\
Keep track of the original question across multiple steps and always make the zoom-in region highly relevant to the question. \\
Only when you are certain about the final answer and no further tool\_call is needed, then output: \\
\texttt{<answer>...</answer>}. \\
\textbf{Format strictly as either:} \\
\texttt{<think>...</think> <tool\_call>...</tool\_call>} \\
\textbf{OR} \\
\texttt{<think>...</think> <answer>...</answer>}
\end{tcolorbox}

\subsection{Clear Prompt}

The prompt explicitly instructs the model to treat zoom-in as a core step in the reasoning process rather than a post-hoc attachment. 
It also constrains the timing of output: if visual evidence is insufficient, the model must zoom in and cannot answer in the same step. 
This structured reasoning format improves both the quality and faithfulness of intermediate steps.

This design ensures that visual inspection is not an afterthought but a structured reasoning action, promoting faithful intermediate steps and robust reasoning under ambiguity.


\begin{figure*}[htbp]
  \centering
  \includegraphics[width=1\linewidth]{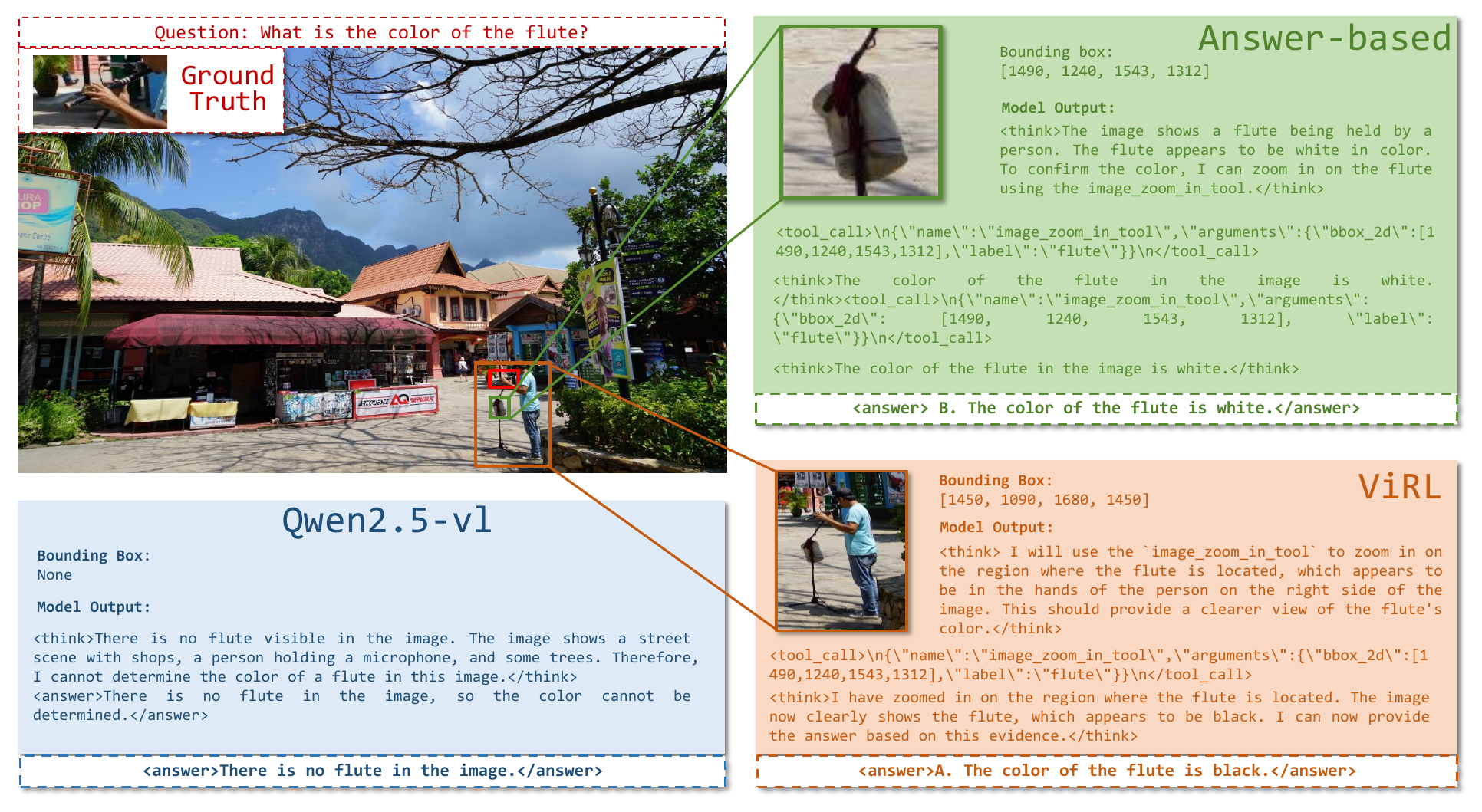}
  \caption{\textbf{Illusion Case.} Wrong visual thinking introduces hallucinatory content, misguiding the model toward an incorrect answer.}
  \label{fig:Illusion2}
\end{figure*}

\begin{figure*}[htbp]
  \centering
  \includegraphics[width=1\linewidth]{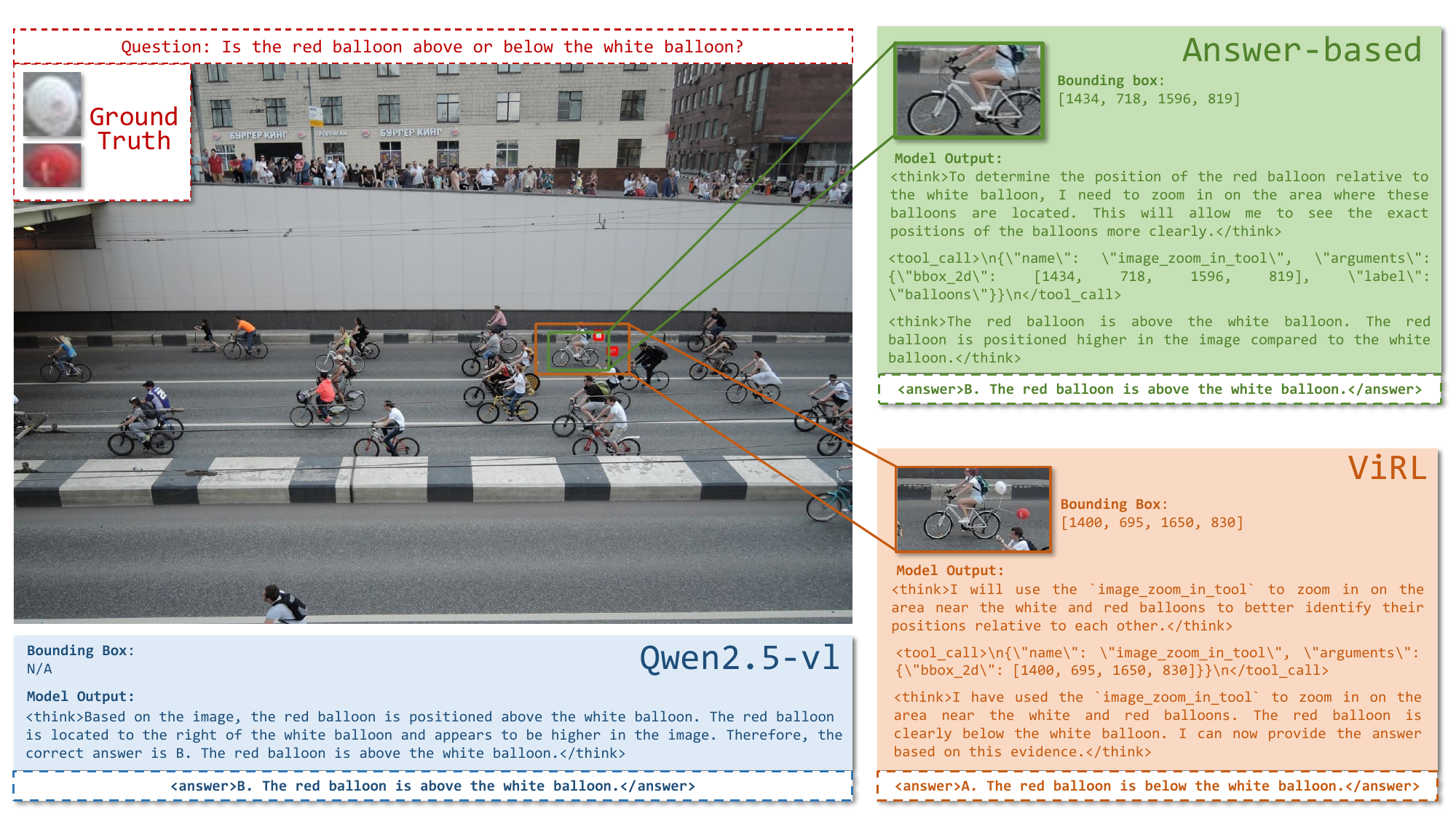}
  \caption{\textbf{Illusion Case.} Incomplete or misleading zoom-in decisions create an illusion of reliable reasoning, causing the model to proceed without valid visual evidence.}
  \label{fig:Illusion3}
\end{figure*}

\begin{figure}[htbp]
  \centering
  \includegraphics[width=1.0\linewidth]{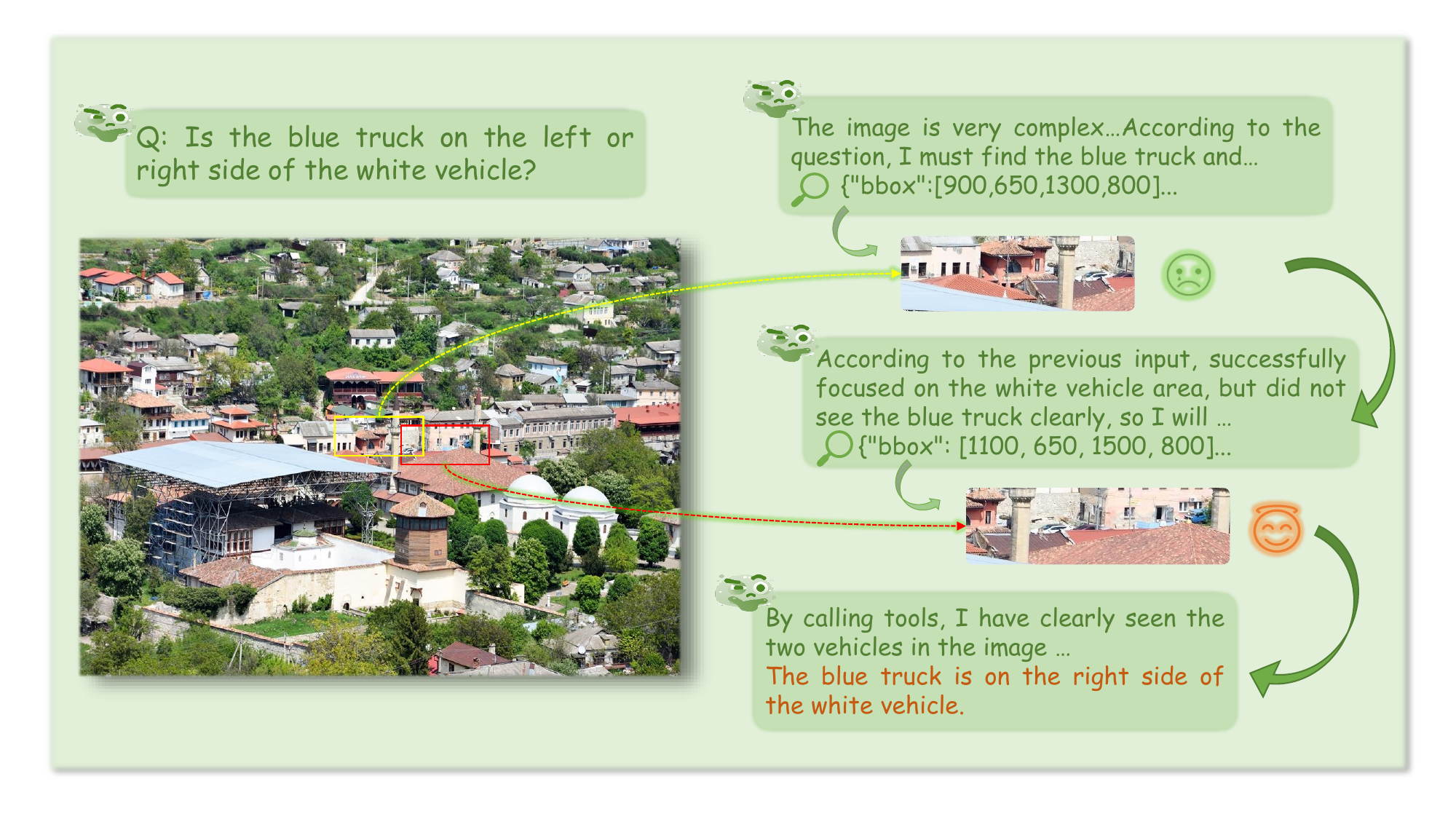}
  \caption{The reasoning process of thinking with images. By embedding zoom in as a structured inference trajectory, more visual cues are provided.}
  \label{fig:inference}
\end{figure}

\begin{figure}[htbp]
  \centering
  \includegraphics[width=0.8\linewidth]{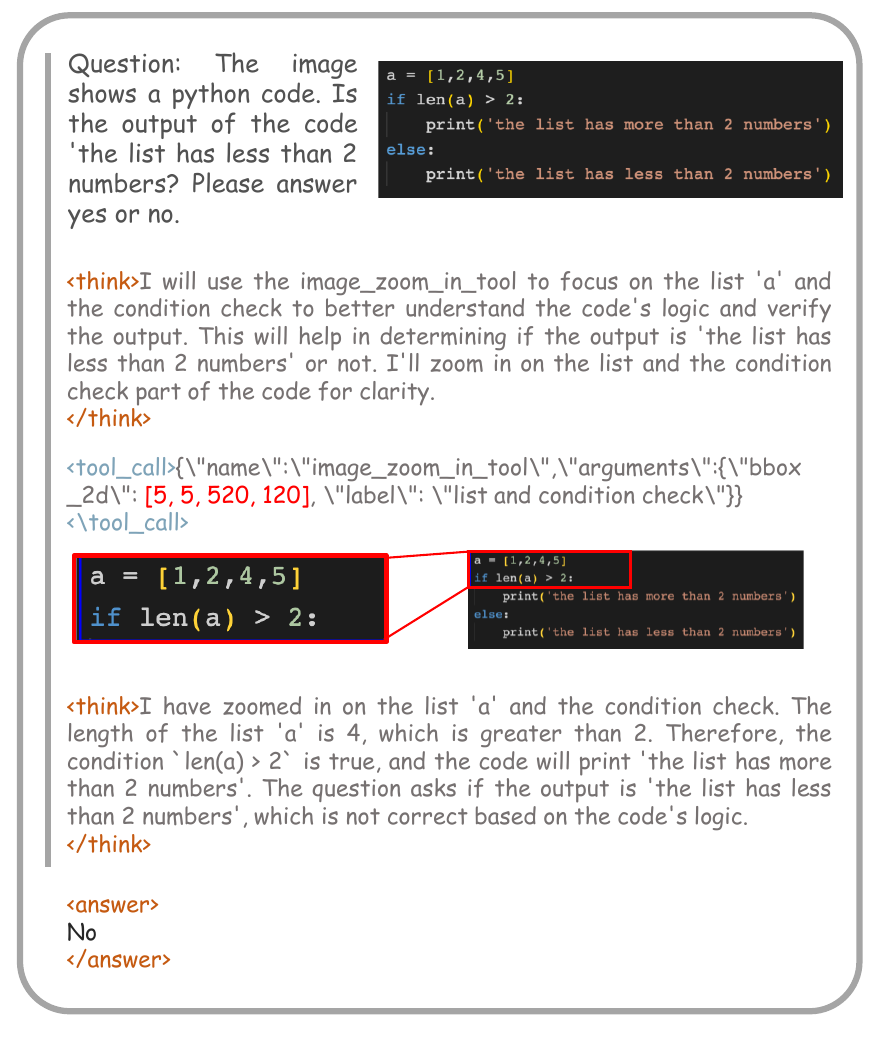}
  \caption{Visual reasoning case1: By embedding zoom in as a structured reasoning form, the traditional COT plain text form is extended, making the reasoning process transparent.}
  \label{fig:reason_case1}
\end{figure}

\section{Qualitative Analysis}
To further illustrate the reasoning challenges and the necessity of visual rationale learning, we present several representative \textbf{illusion cases} observed during inference.
These cases highlight different types of failures in the visual reasoning process that cannot be captured by answer accuracy alone. 
By examining the intermediate thinking steps and invoked visual regions, we can better understand how reasoning illusions emerge and why they undermine trustworthiness.

Specifically, we categorize illusion cases into these types:

\begin{itemize}
    \item \textbf{Hallucinatory Visual Thinking (Error):} The model zooms into irrelevant or non-existent regions, introducing hallucinated visual content that leads to an incorrect answer. As seen in Figures~\ref{fig:Illusion2}.
    This reflects a direct breakdown in perception–reasoning alignment.
    \item \textbf{Hallucinatory Visual Thinking (Hidden Risk):} The model makes relevant but incomplete zoom-in decisions. This creates a dangerous illusion of reliability, as the output is not grounded in valid evidence, and the model does not reason further based on the incomplete clues. As seen in Figures~\ref{fig:Illusion3}.
\end{itemize}

Overall, these cases reveal that visual thinking quality is not strictly coupled with answer accuracy: models can exhibit reasoning illusions that degrade interpretability, efficiency, and safety. 
Therefore, evaluating and improving the faithfulness of visual reasoning trajectories is essential for building trustworthy multimodal systems.

To further illustrate how visual rationale learning enables genuine integration of perception into reasoning, we provide an additional qualitative example in Fig.~\ref{fig:inference}.
Rather than merely triggering a tool call, the model learns to iteratively refine its visual inspection based on previous feedback. When an initial visual attempt yields ambiguous or incorrect cues, the model performs a second round of targeted visual reasoning, progressively converging toward stable and reliable evidence that grounds the final answer.
This adaptive refinement highlights a fundamental shift from shallow tool invocation to feedback-driven visual cognition.

\begin{figure}[htbp]
  \centering
  \includegraphics[width=0.8\linewidth]{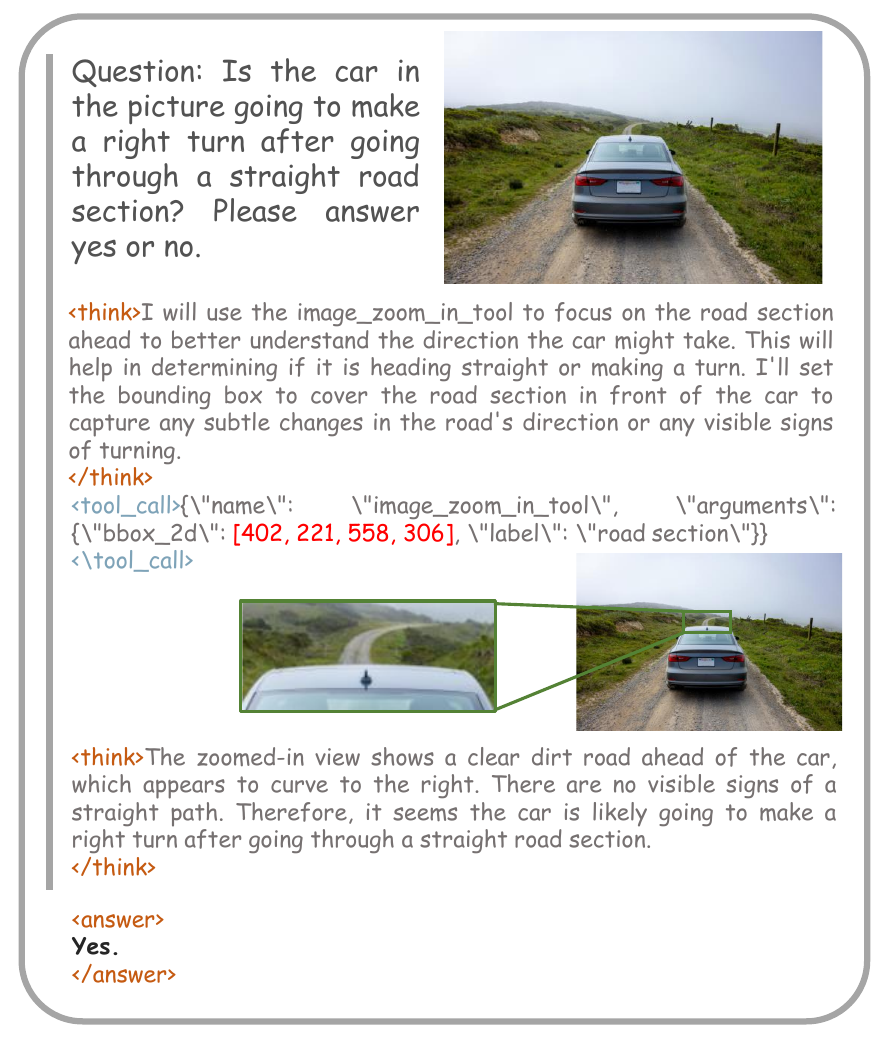}
  \caption{Visual reasoning case2: By embedding zoom in as a structured reasoning form, the traditional COT plain text form is extended, making the reasoning process transparent.}
  \label{fig:reason_case2}
\end{figure}


\begin{figure}[!t]
  \centering
  \includegraphics[width=0.8\linewidth]{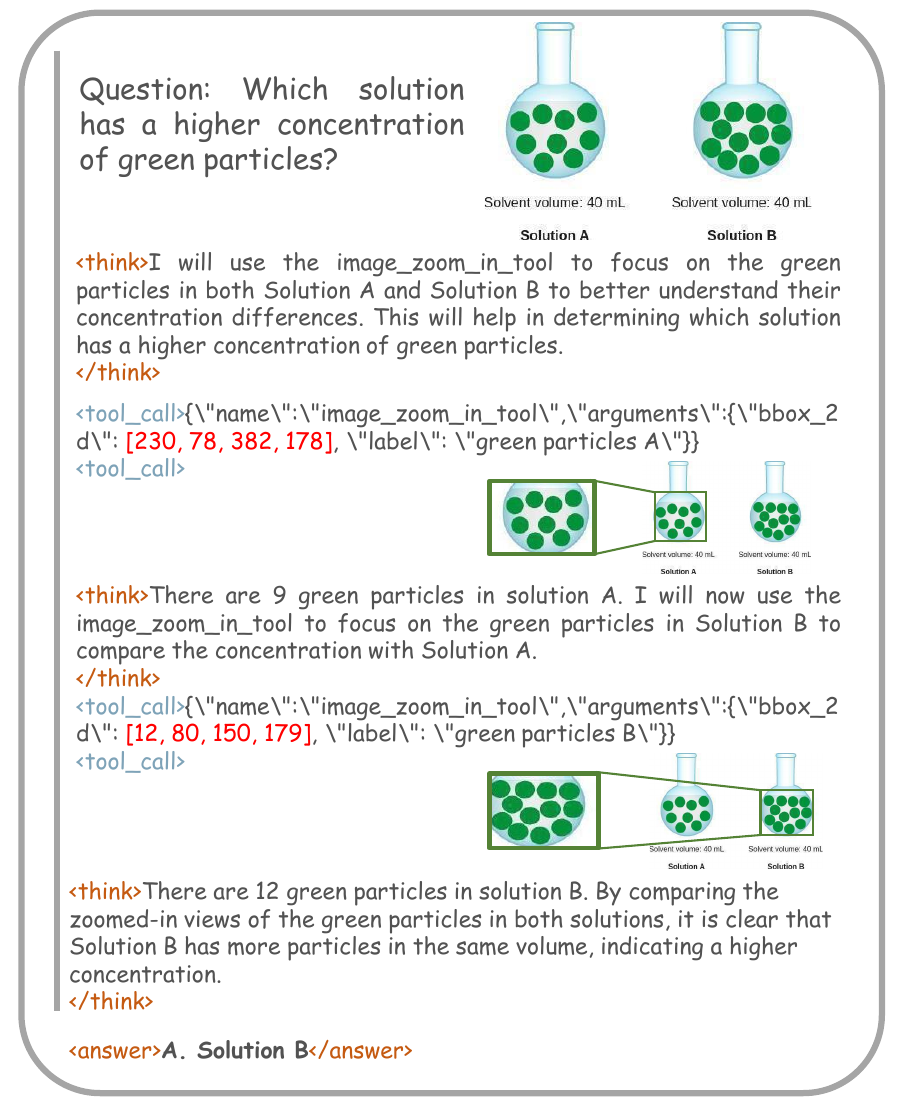}
  \caption{Visual reasoning case3: By embedding zoom in as a structured reasoning form, the traditional COT plain text form is extended, making the reasoning process transparent.}
  \label{fig:reason_case3}
\end{figure}

While our main experiments focus on perception-centric tasks, \textbf{ViRL} also demonstrates strong generalization to broader reasoning scenarios. As shown in Fig.~\ref{fig:reason_case1},~\ref{fig:reason_case2},~\ref{fig:reason_case3}, the model leverages visual rationale not only for fine-grained perceptual inspection but also as an integral part of \emph{general reasoning}.
In these cases, visual thinking serves a similar role to textual chain-of-thought—enabling the model to extract and verify visual evidence step by step to support its reasoning trajectory.
This grounding process enhances the logical soundness and trustworthiness of the model’s outputs, reduces hallucinations, and makes intermediate reasoning steps more interpretable and verifiable.

\begin{figure}[htbp]
  \centering
  \includegraphics[width=1.0\linewidth]{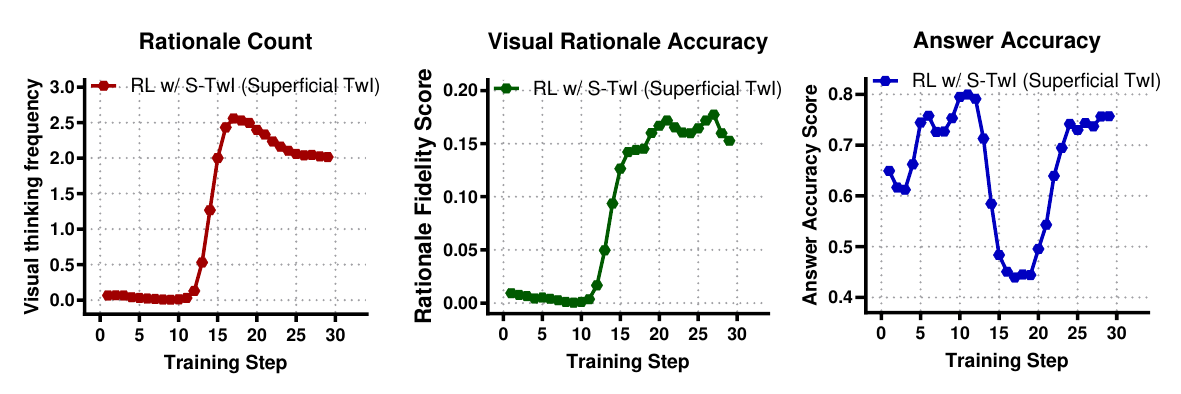}
  \caption{\textbf{Training explosion of zoom-in activity.} A rapid surge in zoom-in calls occurs during early training, leading to a sharp drop in answer accuracy. 
    This explosion phase reflects ungrounded visual reasoning behavior amplified by naive zoom-in rewards, highlighting the need for more structured and discriminative reward design.}
  \label{fig:Training_curve}
\end{figure}

\begin{figure*}[htbp]
  \centering
  \includegraphics[width=1.0\linewidth]{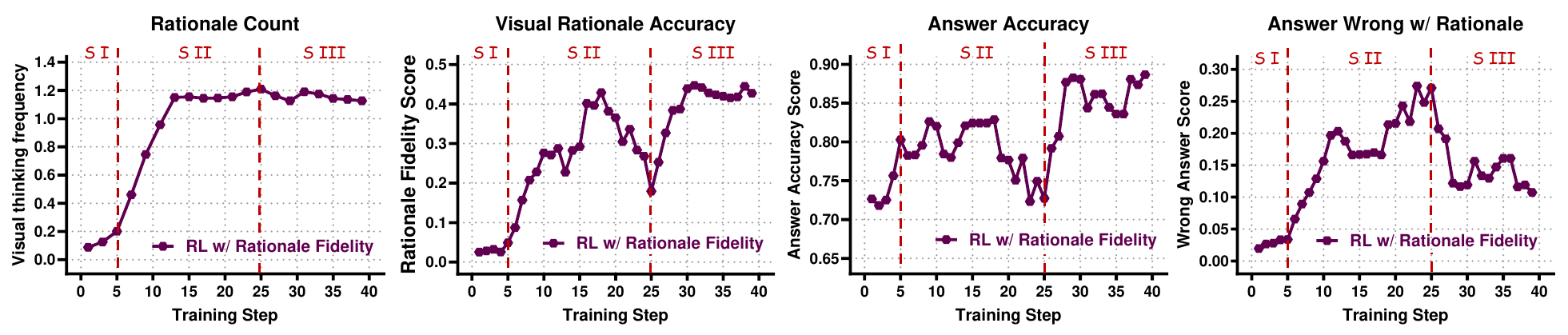}
  \caption{The Dynamic Learning Process of Visual Rationale Learning. Via ViRL, the model transitions from early-stage unfamiliar visual thinking to blind exploration, and finally to confident evidence-based reasoning—progressively integrating visual grounding into the core of its multi-turn reasoning process.}
  \label{fig:inght2}
\end{figure*}

\section{Training Curves}

\paragraph{Reward Hacking.}
As shown in Figures~\ref{fig:Training_curve}, a distinctive pattern emerges in the training: a sharp \emph{explosion of visual thinking activity}, where the frequency of zoom-in calls rapidly increases within a short interval. 
However, this surge is not accompanied by improved reasoning quality. On the contrary, due to immature visual reasoning skills, the model introduces a large amount of irrelevant or noisy visual evidence, causing a steep drop in answer accuracy. 

This phenomenon reveals a critical negative side effect of naively rewarding zoom-in behavior: the model overuses the tool without meaningful grounding, amplifying reasoning illusions and destabilizing early learning.
Although answer accuracy eventually recovers as the model learns to self-regulate its zoom-in usage, this explosion phase reflects a deeper issue closely tied to the illusion of thinking with images: the model learns to perform visual reasoning behaviorally without genuinely integrating visual evidence into its decision process.
Such outcome-driven dynamics encourage superficial exploration rather than structured reasoning, leading to suboptimal convergence and unstable training—ultimately exposing the gap between apparent tool usage and truly grounded reasoning.

\paragraph{The staged dynamics of Visual Rationale Learning.}
Our experiments reveal that robust ``thinking with images'' is not an automatic by-product of answer learning but a staged process (Fig.~\ref{fig:inght2}):
\begin{itemize}[leftmargin=1.5em]
\item \textbf{Stage I — Answer-first learning.} The agent initially optimizes for surface correlations to produce correct answers; zoom-in attempts are noisy and yield little benefit, so the model learns to avoid them.
\item \textbf{Stage II — Ineffective Exploration \& Performance Dip.} The agent begins to explore the tool at a high frequency, but its actions are often erroneous or misaligned. This injects spurious evidence and creates a strong interference, paradoxically causing a significant dip in answer accuracy and a spike in tool-related errors. Crucially, the agent does not become tool-avoidant but persists through this period of intense but inefficient exploration.
\item \textbf{Stage III — visual thinking stabilization.} With dense, per-action feedback and appropriate priors, useful invocations are reinforced and stabilized; thereafter, zoom-ins and answer learning enter a mutually reinforcing regime where visual actions meaningfully improve performance.
\end{itemize}
This staged view explains why answer competence alone does not imply image-grounded reasoning: learners must be guided through exploration and selective reinforcement before visual actions become net-positive.

\section{Limitations and Future Work}

Although our approach demonstrates clear benefits in activating structured visual reasoning, it still faces several limitations.  
First, the dataset remains relatively small and primarily focuses on perception-driven reasoning tasks, limiting its coverage of more complex, multi-hop reasoning scenarios.  
Second, current supervision mainly targets short reasoning chains; extending it to deeper, longer-horizon reasoning remains an open challenge.  

In future work, we plan to expand the dataset to support richer and more diverse reasoning patterns, including multi-step causal inference and math reasoning. We also aim to develop more scalable supervision strategies to broaden the applicability of process-grounded training without sacrificing data quality.

 \fi

\end{document}